\PassOptionsToPackage{numbers}{natbib}
\documentclass{article}

\usepackage[final]{neurips_2020}

\usepackage[utf8]{inputenc} % allow utf-8 input
\usepackage[T1]{fontenc}    % use 8-bit T1 fonts
\usepackage{hyperref}       % hyperlinks
\usepackage{url}            % simple URL typesetting
\usepackage{booktabs}       % professional-quality tables
\usepackage{amsfonts}       % blackboard math symbols
\usepackage{amsmath,amsthm,amssymb,bm}
\usepackage{nicefrac}       % compact symbols for 1/2, etc.
\usepackage{microtype}      % microtypography
\usepackage{graphicx,caption}
\usepackage{epstopdf}
\usepackage{booktabs}
\usepackage{multirow}
\usepackage{wrapfig}
\usepackage{comment}
\usepackage{xr-hyper}
\usepackage{pdfpages}

\title{Temporal Spike Sequence Learning via Backpropagation for Deep Spiking Neural Networks}

\author{
  Wenrui Zhang\\
  University of California, Santa Barbara\\
  Santa Barbara, CA 93106 \\
  \texttt{wenruizhang@ucsb.edu} \\
  \And
  Peng Li\\
  University of California, Santa Barbara\\
  Santa Barbara, CA 93106 \\
  \texttt{lip@ucsb.edu} \\
}

\begin{document}

\maketitle

\begin{abstract}
Spiking neural networks (SNNs) are well suited for spatio-temporal learning and implementations on energy-efficient event-driven neuromorphic processors. However, existing SNN error backpropagation (BP)  methods lack proper handling of spiking discontinuities and suffer from low performance compared with the BP methods for traditional artificial neural networks. In addition, a large number of time steps are typically required to achieve decent performance, leading to high latency and rendering spike based computation unscalable to deep architectures. We present a novel Temporal Spike Sequence Learning Backpropagation (TSSL-BP) method for training deep SNNs, which breaks down error backpropagation across two types of inter-neuron and intra-neuron dependencies and leads to improved temporal learning precision.  It captures inter-neuron dependencies through presynaptic firing times by considering the all-or-none characteristics of firing activities, and captures intra-neuron dependencies by handling the internal evolution of each neuronal state in time.   TSSL-BP efficiently trains deep SNNs within a much  shortened temporal window of a few steps while improving the accuracy for various image classification datasets including CIFAR10. 

%including achieving up to $4\%$ accuracy improvement over the previously reported SNN work on CIFAR10.
\end{abstract}

\section{Introduction}
Spiking neural networks (SNNs), a brain-inspired computational model, have gathered significant interests~\cite{gerstner2002spiking}. 
%Compared with the conventional artificial neural networks (ANNs), 
%that are characterized by a single, static, and continuous-valued activation function
%SNNs are well suitpatio-temporal patterns~\cite{gerstner2002spiking}. 
%
The spike-based operational principles of SNNs not only allow  information coding based on  efficient temporal codes and give rise to promising spatiotemporal computing power but also render energy-efficient VLSI neuromorphic chips such as IBM's TrueNorth~\cite{akopyan2015truenorth} and Intel's Loihi~\cite{davies2018loihi}. Despite the recent progress in SNNs and neuromorphic processor designs, fully leveraging the theoretical computing advantages of SNNs over traditional artificial neural networks (ANNs)~\cite{maass1997networks} to achieve competitive performance in wide ranges of challenging real-world tasks remains difficult. 

Inspired by the success of error backpropagation (BP) and its variants in training conventional deep neural networks (DNNs), various SNNs BP methods have emerged, aiming at attaining the same level of performance~\cite{bohte2002error, lee2016training, wu2018spatio, jin2018hybrid, shrestha2018slayer, zenke2018superspike, huh2018gradient, bellec2018long, bellec2019biologically, zhang2019spike}. Although many appealing results are achieved by these methods, developing SNNs BP training methods that are on a par with the mature BP  tools widely available for training ANNs today is a nontrivial problem~\cite{tavanaei2018deep}. 

Training of SNNs via BP are challenged by two fundamental issues. First, from an algorithmic perspective, the complex neural dynamics
in both spatial and temporal domains make the BP process obscure. Moreover, the errors are hard to be precisely backpropagated due to the non-differentiability of discrete spike events. Second, a large number of time steps are typically required for emulating SNNs in time to achieve decent performance, leading to high latency and rendering spike based computation unscalable to deep architectures. It is desirable to demonstrate the success of BP in training  deeper  SNNs achieving satisfactory performances on more challenging datasets.   
We propose a new SNNs BP method, called temporal spike sequence learning via BP (TSSL-BP), to learn any target output temporal spiking sequences. TSSL-BP acts as a universal training method for any employed spike codes (rate, temporal, and combinations thereof). To tackle  the above difficulties, TSSL-BP breaks down error backpropagation across two types of inter-neuron and intra-neuron dependencies, leading to improved temporal learning precision.  It captures the inter-neuron dependencies within an SNN by considering the all-or-none characteristics of firing activities through presynaptic firing times; it considers the internal evolution of each neuronal state through time, capturing how the  intra-neuron dependencies between different firing times of the same presynaptic neuron impact its postsynaptic neuron. The efficacy and precision of TSSL-BP allows it  to successfully train SNNs over a very short temporal window, e.g. over 5-10 time steps,  enabling ultra-low latency spike computation. As shown in Section~ \ref{sec:results}, 
TSSL-BP signficantly improves accuracy and runtime efficiency of BP training on several well-known image datasets of MNIST~\cite{lecun1998gradient}, NMNIST~\cite{orchard2015converting}, FashionMNIST~\cite{xiao2017fashion}, and CIFAR10~\cite{krizhevsky2014cifar}. Specifically, it achieves up to $3.98\%$ accuracy improvement over the previously reported SNN work on CIFAR10, a challenging dataset for all prior SNNs BP methods.

\section{Background}

\subsection{Existing Backpropagation methods for SNNs}
% Backpropagation (BP) is the workhorse for training deep ANNs~\cite{lecun2015deep}. Its success in the ANN world has made BP a target of intensive research for SNNs. Nevertheless, applying BP to biologically more plausible SNNs is nontrivial due to the necessity in dealing with complex neural dynamics and non-differentiability of discrete spike events. 

One of the earliest SNNs BP methods is the well-known SpikeProp algorithm~\cite{bohte2002error}. However, SpikeProp and its variants~\cite{booij2005gradient, ghosh2009new, xu2013supervised} are still limited to single spike per output neuron without demonstrating competitive performance on real-world tasks. In addition, \cite{diehl2015fast, esser2015backpropagation, hunsberger2016training, sengupta2019going} train ANNs and then approximately convert them to SNNs. Nevertheless, such conversion leads to approximation errors and cannot exploit SNNs' temporal learning capability.

Recently, training SNNs with BP under a firing rate (or activity level) coded loss function has been shown to deliver competitive performances~\cite{lee2016training, wu2018spatio, shrestha2018slayer,  jin2018hybrid, zenke2018superspike, huh2018gradient, bellec2018long, zhang2019spike}. Among them, \cite{lee2016training} does not consider the temporal correlations of neural activities and treats spiking times as noise to allow error gradient computation.  \cite{shrestha2018slayer, wu2018spatio, zenke2018superspike, bellec2018long} capture the temporal effects by performing backpropagation through time (BPTT)~\cite{werbos1990backpropagation}. 
However, they get around the non-differentiability of spike events by approximating the spiking process using the surrogate gradient method~\cite{neftci2019surrogate}. These approximations lead to inconsistency between the computed gradient and target loss, and thus degrade training performance. 
\cite{huh2018gradient} presents a BP method for recurrent SNNs based on a novel combination of a gate function and threshold-triggered synaptic model that are introduced to handle non-differentiability of spikes. In this work, depolarization of membrane potential within a narrow active zone below the firing threshold also induces graded postsynaptic current. 
\cite{jin2018hybrid, zhang2019spike} present spike-train level BP methods by capturing the effects of spiking neurons aggregated at the spike train level. However, the length of spike trains over which BP is applied need to be long enough, leading to long inference latency and high training cost.

\cite{wu2019direct} can train SNNs over a relatively small number of time steps by adding optimization techniques such as neuron normalization and population decoding. Since its core lies at the method of \cite{wu2018spatio}, it still approximates the all-or-none firing characteristics of spiking neurons by a continuous activation function, causing the same problems introduced before.

In this work, we propose the TSSL-BP method as a universal training method for any employed spike codes. It can not only precisely capture the temporal dependencies but also allow ultra-low latency inference and training over only five time steps while achieving excellent accuracies. 

\subsection{Spiking Neuron Model}
% introduction of LIF model
SNNs employ more biologically plausible spiking neuron models than ANNs. In this work, we adopt the leaky integrate-and-fire (LIF) neuron model and synaptic model \cite{gerstner2002spiking}.

Consider the input spike train from pre-synaptic neuron $j$ to neuron $i$: $s_{j}(t)=\sum_{t_j^{(f)}}\delta(t-t_j^{(f)})$, where $t_j^{(f)}$ denotes a particular firing time of presynaptic neuron $j$. The incoming spikes are converted into an (unweighted) postsynaptic current (PSC) $a_j(t)$ through a synaptic model. The neuronal membrane voltage $u_i(t)$ at time $t$ for neuron $i$ is given by
\begin{equation}
    \label{eq:lif}
    \tau_m \frac{du_i(t)}{dt} = -u_i(t) + R~\sum_j w_{ij}a_j (t) + \eta_i(t), 
\end{equation}
where $R$ and $\tau_m$ are the effective leaky resistance and time constant of the membrane, $w_{ij}$ is the synaptic weight  from  pre-synaptic neuron $j$ to neuron $i$, $a_j(t)$ is the (unweighted) postsynaptic potential (PSC) induced by the spikes from pre-synaptic neuron $j$, and $\eta(t)$ denotes the reset function. 

The PSC and the reset function can be written as
\begin{equation}
    \label{eq:psc_reset}
        a_j(t)=(\epsilon*s_j)(t), \qquad \eta_i(t)=(\nu*s_i)(t),
\end{equation}
where $\epsilon(\cdot)$ and $\nu(\cdot)$ are the spike response and  reset kernel, respectively. In this work, we adopt a first order synaptic model as the spike response kernel which is expressed as:
\begin{equation}
    \label{eq:first-order}
    \tau_s \frac{a_j(t)}{dt} = -a_j(t) + s_{j}(t),
\end{equation}
where $\tau_s$ is the synaptic time constant.

The reset kernel reduces the membrane potential by a certain amount $\Delta_R$, where $\Delta_R$ is equal to the firing threshold right after the neuron fires. %and then decays over time according to the time constant of the membrane potential $\tau_m$. 
Considering the discrete time steps simulation, we use the fixed-step first-order Euler method to discretize (\ref{eq:lif}) to
\begin{equation}
    \label{eq:discrete_lif}
    u_i[t] = (1 - \frac{1}{\tau_m}) u_i[t-1] + \sum_j w_{ij}a_j [t] + \eta_i[t].
\end{equation}
The ratio of R and $\tau_m$ is absorbed into the synaptic weight. The reset function $\eta_i[t]$ represents the firing-and-resetting mechanism of the neuron model. Moreover, the firing output of the neuron is expressed as
\begin{equation}
    \label{eq:fire_function}
    s_i[t] = H\left(u_i[t] - V_{th}\right),
\end{equation}
where $V_{th}$ is the firing threshold and $H(\cdot)$ is the Heaviside step function. %with the form
%\begin{equation}
%    H(p)=\begin{cases}
%    0 & p<0\\
%    1 & p\geq0
%\end{cases}.
%\end{equation}

%In the following section, we mainly use (\ref{eq:discrete_lif}) and (\ref{eq:fire_function}) to derive the proposed TSSL-BP method.

\section{Methods}
\subsection{Forward Pass}
\begin{wrapfigure}{r}{0.5\textwidth}
    \centering
        \includegraphics[width=0.5\textwidth]{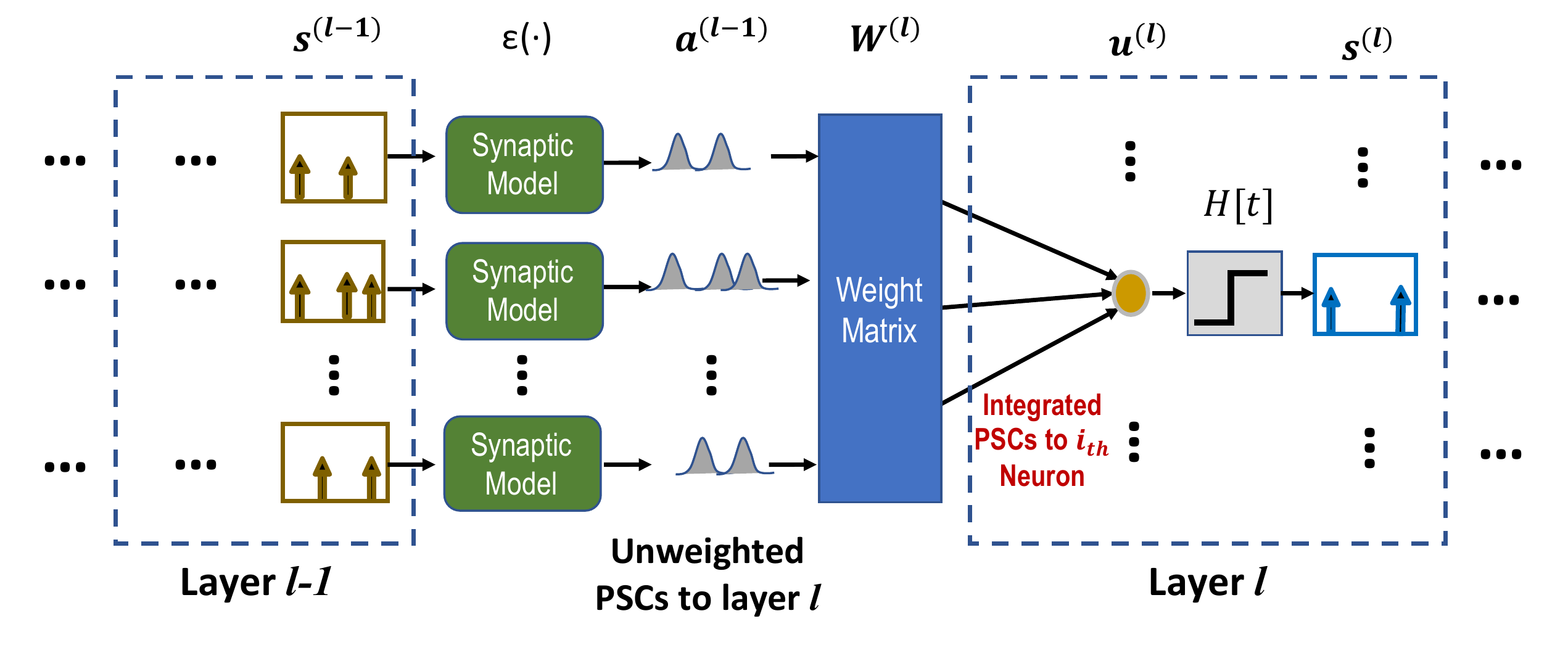}
    \centering
    \captionsetup{width=.9\linewidth}
    \caption{Forward evaluation pass of SNNs.}
    \label{fig:tssl_bp_forward_backward_pass}
\end{wrapfigure} 
Without loss of generality, we consider performing BP across two adjacent layers $l-1$ and $l$ with $N_{l-1}$ and $N_{l}$ neurons, respectively, in a fully-connected feedforward SNNs as shown in Figure \ref{fig:tssl_bp_forward_backward_pass}. The procedure can be also applied to convolutional and pooling layers. Denote the presynaptic weights by $\bm{W}^{(l)}=\left[\bm{w}^{(l)}_1, \cdots, \bm{w}^{(l)}_{N_{l}}\right]^T$, where $\bm{w}^{(l)}_{i}$ is a column vector of weights from all the neurons in layer $l-1$ to the neuron $i$ of layer $l$. In addition, we also denote PSCs from neurons in layer $l-1$  by   $\bm{a}^{(l-1)}[t]=\left[a^{(l-1)}_{1}[t], \cdots, a^{(l-1)}_{N_{l-1}}[t]\right]^T$, spike trains output of the $l-1$ layer by $\bm{s}^{(l-1)}[t]=\left[s^{(l-1)}_{1}[t], \cdots, s^{(l-1)}_{N_{l-1}}[t]\right]^T$, membrane potentials and the corresponding output spike trains of the $l$ layer neurons respectively by $\bm{u}^{(l)}[t]=\left[u^{(l)}_{1}[t], \cdots, u^{(l)}_{N_{l}}[t]\right]^T$ and  $\bm{s}^{(l)}[t]=\left[s^{(l)}_{1}[t], \cdots, s^{(l)}_{N_{l}}[t]\right]^T$, where variables associated with neurons in the layer $l$ have $l$ as the superscript.

The forward propagation between the two layers is described as
\small
\begin{align}
    \label{eq:forward_pass}
    \begin{split}
        & \bm{a}^{(l-1)}[t] = (\epsilon * \bm{s}^{(l-1)})[t], \quad \bm{u}^{(l)}[t] = (1 - \frac{1}{\tau_m}) \bm{u}^{(l)}[t-1] + \bm{W}^{(l)}\bm{a}^{(l-1)}[t] + (\nu * \bm{s}^{(l)})[t], \\
        & \bm{s}^{(l)}[t] = H\left(\bm{u}^{(l)}[t] - V_{th}\right).
    \end{split}
\end{align}
\normalsize
In the forward pass, the spike trains $\bm{s}^{(l-1)}[t]$ of the $l-1$ layer generate the (unweighted) PSCs $\bm{a}^{(l-1)}[t]$ according to the synaptic model. Then, $\bm{a}^{(l-1)}[t]$ are multiplied the synaptic weights and passed onto the neurons of layer $l$. The integrated PSCs alter the membrane potentials and trigger the output spikes of the layer $l$ neurons when the membrane potentials exceed the threshold.
% \begin{figure}[ht]
% \begin{center}
% \includegraphics[width=0.9\textwidth]{Figure/forward_pass.pdf}
% \end{center}
% \caption{Forward evaluation pass of SNNs.}
% \label{fig:tssl_bp_forward_backward_pass}
% \end{figure}

\subsection{The Loss Function}
The goal of the proposed TSSL-BP method is to train a given SNN in such a way that each output neuron learns to produce a desired firing sequence arbitrarily specified by the user  according to the input class label.  Denote the desired and the actual spike trains in the output layer by $\bm{d} = \left[\bm{d}[t_0], \cdots, \bm{d}[t_{N_t}]\right]$ and $\bm{s} = \left[\bm{s}[t_0], \cdots, \bm{s}[t_{N_t}]\right]$ where $N_t$ is the number of the considered time steps, $\bm{d}[t]$ and $\bm{s}[t]$ the desired and actual firing events for all output neurons at time $t$, respectively. 

The loss function $L$ can be defined using any suitable  distance function  measuring the difference between $\bm{d}$ and $\bm{s}$. In this work, the loss function is defined by the total square error summed over each output neuron at each time step:
\begin{equation}
\label{eq:loss}
        L = \sum_{k=0}^{N_t}E[t_k] = \sum_{k=0}^{N_t}\frac{1}{2}((\epsilon * \bm{d})[t_k] - (\epsilon * \bm{s})[t_k])^2,
\end{equation}
where $E[t]$ is the error at time $t$ and $\epsilon(\cdot)$ a kernel function which measures the so-called Van Rossum distance between the actual spike train and desired spike train. 

% The loss can be also defined by using a kernel function \cite{shrestha2018slayer}. Using the spike response kernel $\epsilon(\cdot)$, one defines the error at each time step as
% \begin{equation}
%     \label{eq:kernel_loss_function}
%         E[t] = (\epsilon * \bm{d})[t] - (\epsilon * \bm{s})[t]=\bm{a_d}[t] - \bm{a_s}[t].
% \end{equation}

\subsection{Temporal Spike Sequence Learning via Backpropagation (TSSL-BP) Method} \label{sec:tssl-bp}

From the loss function (\ref{eq:loss}), we define the error $E[t_k]$ at each time step. $E[t_k]$ is based on the output layer firing spikes at $t_k$ which further depend on all neuron states ${\bf u}[t]$,  $t \leq t_k$. Below we consider these temporal dependencies and derive the main steps of the proposed TSSL-BP method. Full details of the derivation are provided in Section 1 of the Supplementary Material.

Using (\ref{eq:loss}) and the chain rule, we obtain
\begin{equation}
    \label{eq:d_l_d_w}
        \frac{\partial L}{\partial \bm{W}^{(l)}} = \sum_{k=0}^{N_t}\sum_{m=0}^{k}\frac{\partial E[t_k]}{\partial \bm{u}^{(l)}[t_m]} \frac{\partial \bm{u}^{(l)}[t_m]}{\partial\bm{W}^{(l)}} = \sum_{m=0}^{N_t}\bm{a}^{(l-1)}[t_m] \sum_{k=m}^{N_t}\frac{\partial E[t_k]}{\partial \bm{u}^{(l)}[t_m]}.   % ToDo transpose
\end{equation}
Similar to the conventional backpropagation, we use $\delta$ to denote the back propagated error at time $t_m$ as $ \bm{\delta}^{(l)}[t_m]=\sum_{k=m}^{N_t}\frac{\partial E[t_k]}{\partial \bm{u}^{(l)}[t_m]}$. $\bm{a}^{(l-1)}[t_m]$ can be easily obtained from (\ref{eq:forward_pass}). $\bm{\delta}^{(l)}[t_m]$ is considered in two cases. % ToDo This statement is not precise

\textbf{[$l$ is the output layer.]} The $\bm{\delta}^{(l)}[t_m]$ can be computed from
\begin{equation}
    \label{eq:delta_output_layer}
    \bm{\delta}^{(l)}[t_m]= \sum_{k=m}^{N_t}\frac{\partial E[t_k]}{\partial \bm{a}^{(l)}[t_k]}\frac{\partial \bm{a}^{(l)}[t_k]}{\partial \bm{u}^{(l)}[t_m]}.
\end{equation}
The first term of (\ref{eq:delta_output_layer}) can be obtained directly from the loss function. The second term $\frac{\partial \bm{a}^{(l)}[t_k]}{\partial \bm{u}^{(l)}[t_m]}$ is the key part of the TSSL-BP method and is discussed in the following sections.

\textbf{[$l$ is a hidden layer.]}  $\bm{\delta}^{(l)}[t_m]$ is derived using the chain rule and (\ref{eq:forward_pass}). 
\small
\begin{align}
    \label{eq:delta_hidden_layer}
    \bm{\delta}^{(l)}[t_m] = \sum_{j=m}^{N_t}\sum_{k=m}^{j}\frac{\partial \bm{a}^{(l)}[t_k]}{\partial \bm{u}^{(l)}[t_m]}\left(\frac{\partial \bm{u}^{(l+1)}[t_k]}{\partial \bm{a}^{(l)}[t_k]}\frac{\partial E[t_j]}{\partial \bm{u}^{(l+1)}[t_k]} \right) = (\bm{W}^{(l+1)})^T\sum_{k=m}^{N_t}\frac{\partial \bm{a}^{(l)}[t_k]}{\partial \bm{u}^{(l)}[t_m]}\bm{\delta}^{(l+1)}[t_k].
\end{align}
\normalsize
(\ref{eq:delta_hidden_layer}) maps the error $\bm{\delta}$ from layer $l+1$ to layer $l$. It is obtained from the fact that membrane potentials $\bm{u}^{(l)}$ of the neurons in layer $l$ influence their (unweighted) corresponding postsynaptic currents (PSCs) $\bm{a}^{(l)}$ through fired spikes, and $\bm{a}^{(l)}$ further affect the membrane potentials $\bm{u}^{(l+1)}$ in the next layer. 

\subsubsection{Key challenges in SNN BackPropagation} 
As shown above, for both the output layer and hidden layers, once $\frac{\partial \bm{a}^{(l)}[t_k]}{\partial \bm{u}^{(l)}[t_m]} (t_k\geq t_m)$ are known, the error $\bm{\delta}$ can be back propagated and the gradient of each layer can be calculated.

Importantly, the dependencies of the PSCs on the corresponding membrane potentials of the presynaptic neurons reflected in $\frac{\partial \bm{a}^{(l)}[t_k]}{\partial \bm{u}^{(l)}[t_m]} (t_k\geq t_m)$  are due to the following spiking neural behaviors:  a change in the membrane potential  may bring it up to the firing threshold, and hence activate the corresponding neuron by generating a spike, which in turn produces a PSC. Computing $\frac{\partial \bm{a}^{(l)}[t_k]}{\partial \bm{u}^{(l)}[t_m]} (t_k\geq t_m)$ involves the activation of each neuron, i.e. firing a spike due to the membrane potential's crossing the firing threshold from below. Unfortunately, the all-or-none firing characteristics of spiking neurons makes the activation function nondifferentiable, introducing several key challenges.  
% \begin{wrapfigure}{r}{0.5\textwidth}
%     \centering
%         \includegraphics[width=0.48\textwidth]{Figure/smooth_function.pdf}
%     \centering
%     \captionsetup{width=.9\linewidth}
%     \caption{“Fictitious” smoothing of activation}
%     \label{fig:smooth_function}
% \end{wrapfigure} 

\begin{figure}[ht]
\centering
\begin{minipage}{.45\textwidth}
  \centering
  \includegraphics[width=.95\linewidth]{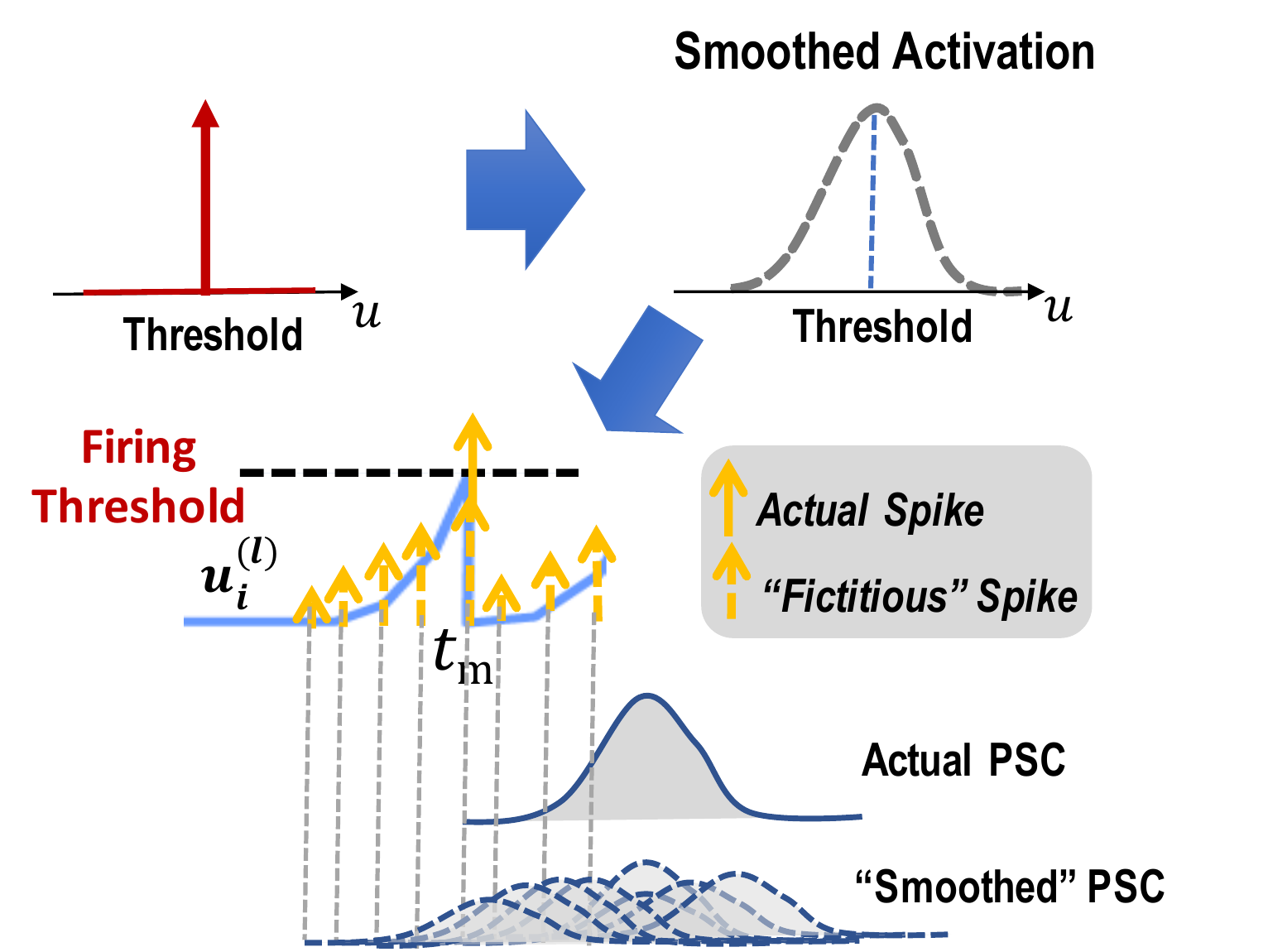}
  \captionof{figure}{“Fictitious” smoothing of activation}
  \label{fig:smooth_function}
\end{minipage}%
\begin{minipage}{.45\textwidth}
  \centering
  \includegraphics[width=.95\linewidth]{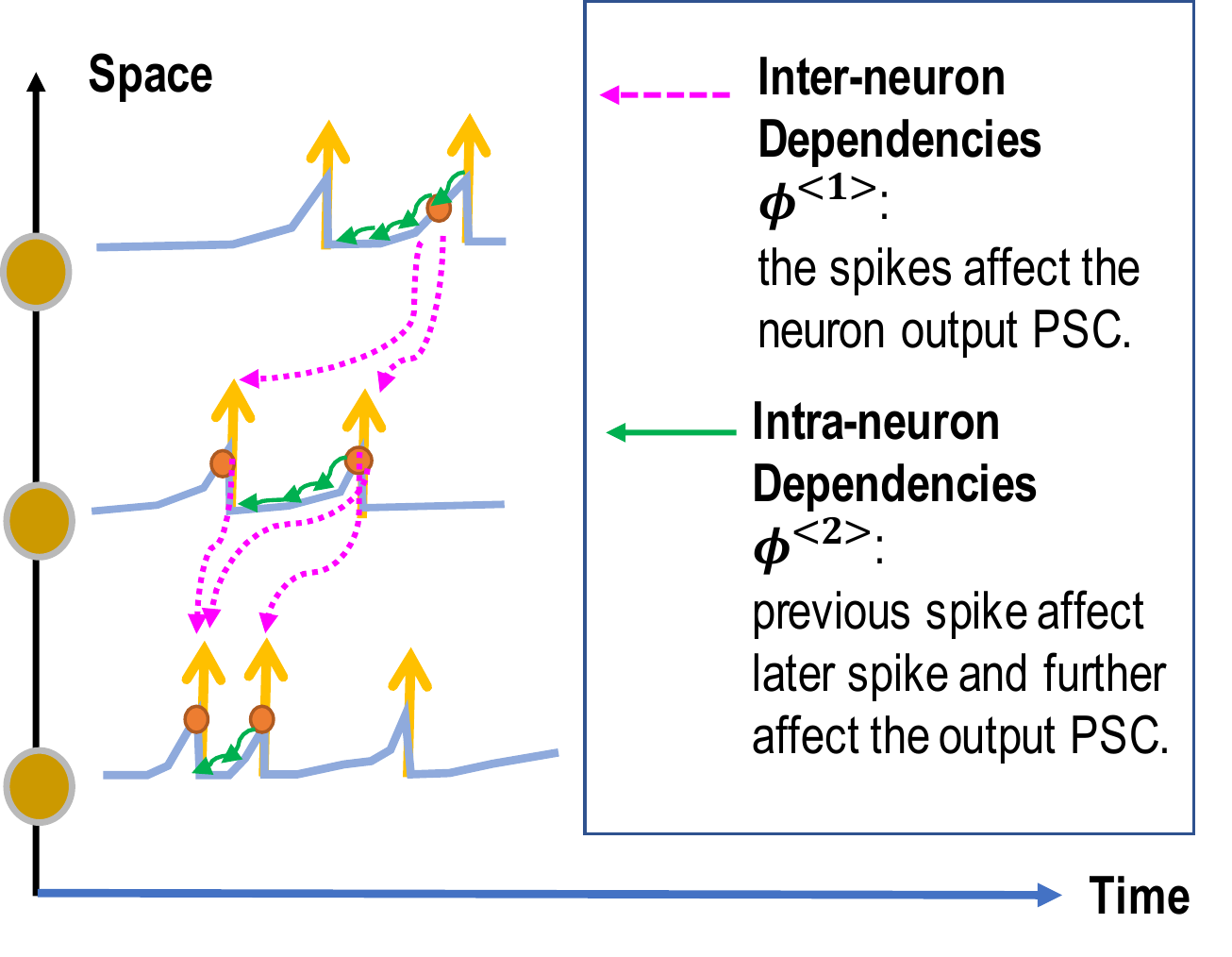}
  \captionof{figure}{Inter/Intra neuron dependencies.}
  \label{fig:inter_intra_dependencies}
\end{minipage}
\end{figure}

A typical strategy in dealing with the non-differentiability of the activation is to smooth the activation function by approximating it using a differentiable curve~\cite{wu2018spatio} as shown in the Figure~ \ref{fig:smooth_function}, or a continuous probability density function~\cite{shrestha2018slayer}, which is similar to the former approach in spirit.   However, these approaches  effectively spread each discrete firing spike continuously over time, converting one actual spike to multiple “fictitious” spikes and also generating multiple “fictitious” PSCs displaced at different time points. We stress that while smoothing circumvents the numerical challenges brought by  non-differentiability of the spiking activation, it effectively alters the underlying spiking neuron model and firing times, and leads to degraded accuracy in the error gradient computation. It is important to reflect that spike timing is the hallmark of spiking neural computation, altering firing times in BP can hamper  precise learning of the targeted firing sequences as pursued in this paper.

\subsubsection{The Main Ideas Behind TSSL-BP}

TSSL-BP addresses the two key limitations of the prior BP methods: lack proper handling of spiking discontinuities (leading to loss of temporal precision) and need for many time steps (i.e. high latency) to ensure good performance.  TSSL-BP computes $\frac{\partial \bm{a}^{(l)}[t_k]}{\partial \bm{u}^{(l)}[t_m]}$ across two categories of spatio-temporal dependencies in the network: \textbf{inter-neuron} and \textbf{intra-neuron} dependencies. As shown in Figure \ref{fig:inter_intra_dependencies}, our key observations are: 1) temporal dependencies of a postsynaptic neuron on any of its presynaptic neurons \emph{only} take place via the presynaptic spikes which generate PSCs to the postsynaptic neuron, and shall be considered as inter-neuron dependencies; 2) furthermore, the timing of one presynaptic spike affects the timing of the immediately succeeding spike from the same presynaptic neuron through the intra-neuron temporal dependency. The timing of the first presynaptic spike affects the PSC produced by the second spike, and has additional impact on the postynaptic neuron through this indirect mechanism.  In the following, we show how to derive the $\frac{\partial a_i^{(l)}[t_k]}{\partial u_i^{(l)}[t_m]}$ for each neuron $i$ in layer $l$. We denote $\phi_i^{(l)}(t_k, t_m) = \frac{\partial a_i^{(l)}[t_k]}{\partial u_i^{(l)}[t_m]} = \phi^{(l)<1>}_i(t_k, t_m) +  \phi^{(l)<2>}_i(t_k, t_m)$, where $\phi^{(l)<1>}_i(t_k, t_m)$ represents the inter-neuron dependencies and $\phi^{(l)<2>}_i(t_k, t_m)$ is the intra-neuron dependencies.

\subsubsection{Inter-Neuron Backpropagation} \label{sec:inter_neuron}
\begin{figure}[ht]
\centering
\begin{minipage}{.6\textwidth}
  \centering
  \includegraphics[width=.95\linewidth]{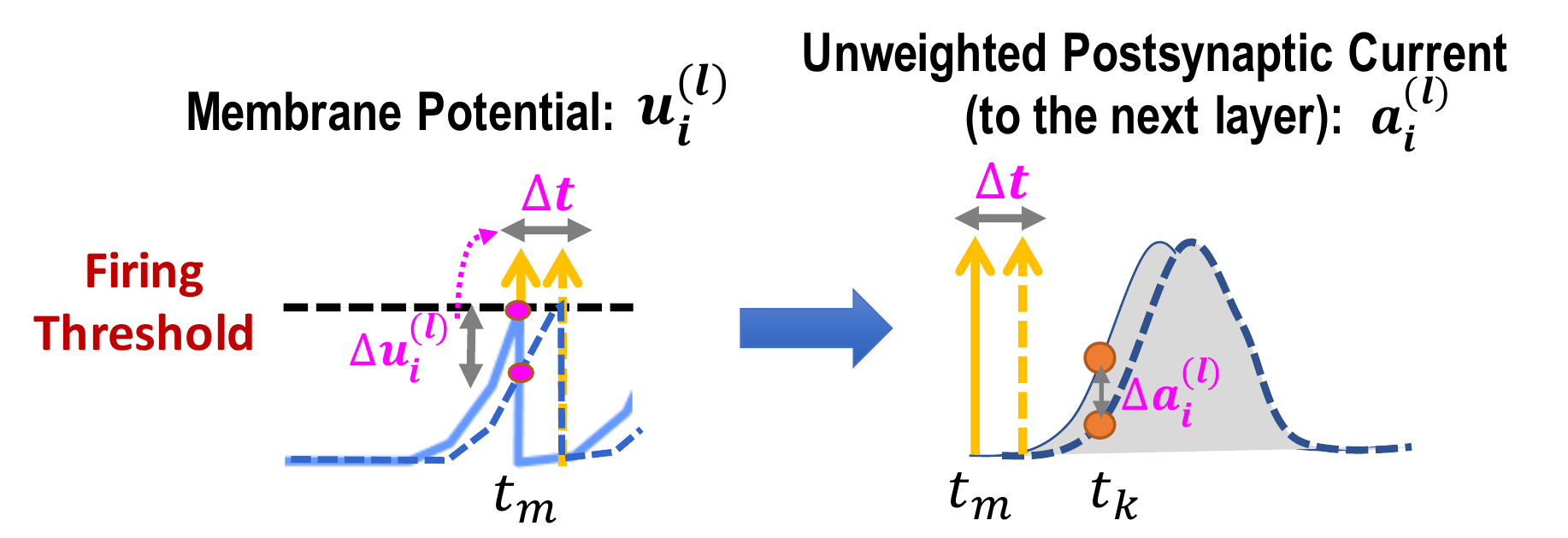}
  \captionof{figure}{PSC dependencies on presynaptic potential.}
  \label{fig:inter_neuron_dependency}
\end{minipage}%
\begin{minipage}{.4\textwidth}
  \centering
  \includegraphics[width=.95\linewidth]{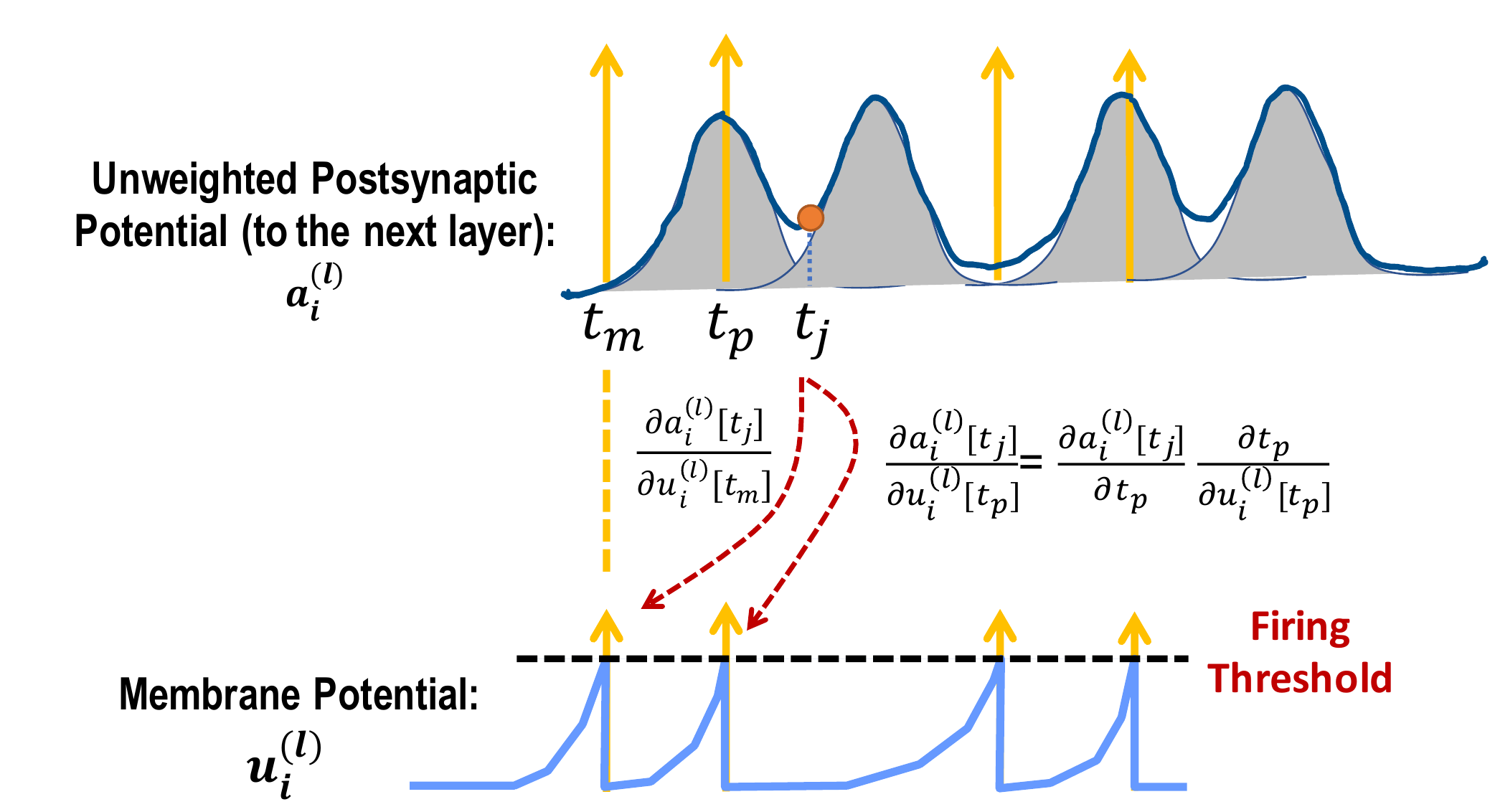}
  \captionof{figure}{Inter-neuron dependencies.}
  \label{fig:derivative_at_firing}
\end{minipage}
\end{figure}

Instead of performing the problematic activation smoothing, we critically note that the all-or-none characteristics of firing behavior is such that a PSC waveform is only triggered at a presynaptic firing time. Specially, as shown in Figure \ref{fig:inter_neuron_dependency}, a perturbation $\Delta u_i^{(l)}$ of $u_i^{(l)}[t_m]$, i.e, due to weight updates, may result in an incremental shift in the firing time $\Delta t$, which in turn shifts the onset of the PSC waveform corresponding to the shifted spike, leading to a perturbation $\Delta a_i^{(l)}$ of $a_i^{(l)}[t_k]$. We consider this as an inter-neuron dependency since the change in PSC ($\Delta a_i^{(l)}$) alters the membrane potential of the postsynaptic neuron in the next layer.

We make two important points: 1) 
%instead of performing ``fictitious" smoothing of the all-or-none spiking activation, 
we shall capture the inter-neuron dependencies via (the incremental changes of) the presynaptic firing times, which precisely corresponds to how different neurons interact with each other in an SNN; and 2) the \emph{inter-neuron} dependency of each neuron $i$’s PSC $a_i^{(l)}$ at $t_k$ on its membrane potential $u_i^{(l)}$ at $t_m$ happens only if the neuron fires at $t_m$. In general, $a_i^{(l)}[t_k]$'s inter-neuron dependencies on \emph{all} preceding firing times shall be considered.  Figure~\ref{fig:derivative_at_firing} shows the situation where $a_i^{(l)}[t_k]$ depends on two presynaptic firing times $t_m$ and $t_p$. Conversely, the inter-neuron dependencies $\phi^{(l)<1>}_i(t_k, t_m) = 0$ if $t_k < t_m$ or there is no spike at $t_m$.
Assuming that the presynaptic neuron $i$ spikes at $t_m$, The inter-neuron dependencies is
\begin{equation}
    \label{eq:inter_neuron_dependency}
        \phi^{(l)<1>}_i(t_k, t_m) 
        % = \frac{\partial a_i^{(l)}[t_k]}{\partial u_i^{(l)}[t_m]} 
        = \frac{\partial a_i^{(l)}[t_k]}{\partial t_m} \frac{\partial t_m}{\partial u_i^{(l)}[t_m]},
\end{equation}
where, importantly, the chain rule is applied through the presynaptic firing time $t_m$.  

From (\ref{eq:psc_reset}), the two parts part of (\ref{eq:inter_neuron_dependency}) can be calculated as
\begin{equation}
\label{eq:dt_du}
        \frac{\partial a_i^{(l)}[t_k]}{\partial t_m} =  \frac{\partial (\epsilon*s^{(l)}_i[t_m])[t_k]}{\partial t_m}, \qquad \frac{\partial t_m}{\partial u_i^{(l)}[t_m]} = \frac{-1}{\frac{\partial u_i^{(l)}[t_m]}{\partial t_m}},
\end{equation}
where $\frac{\partial u_i^{(l)}[t_m]}{\partial t_m}$ is obtained by differentiating (\ref{eq:discrete_lif}).

% The term $\frac{\partial t_m}{\partial u_i^{(l)}[t_m]}$ cannot be computed directly because time cannot be expressed as a differentiable function of membrane potential. \cite{bohte2002error} proposed a solution in the SpikeProp method by the assuming that $u_i(t)$ is a linear function of $t$ around the output spike time $t = t_i^{(f)}$. In addition, \cite{ghosh2009new} used the same solution but offered a clearer explanation. Therefore, by adopting the same solution, the second part of (\ref{eq:inter_neuron_dependency}) can be expressed as
% \begin{equation} \label{eq:dt_du}
%         \frac{\partial t_m}{\partial u_i^{(l)}[t_m]} = \frac{-1}{\frac{\partial u_i^{(l)}[t_m]}{\partial t_m}},
% \end{equation}
% where $\frac{\partial u_i^{(l)}[t_m]}{\partial t_m}$ can be easily obtained from (\ref{eq:discrete_lif}).

\subsubsection{Intra-Neuron Backpropagation}\label{sec:intra_neuron}

\begin{wrapfigure}{r}{0.35\textwidth}
    \centering
        \includegraphics[width=0.35\textwidth]{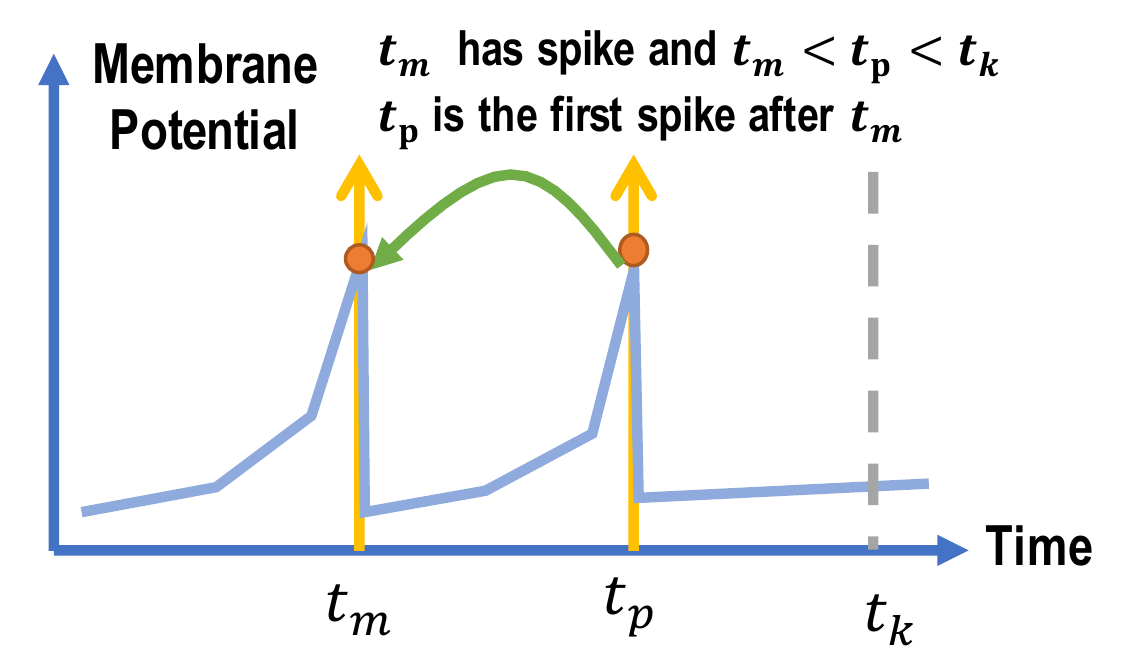}
    \centering
    \captionsetup{width=.9\linewidth}
    \caption{Intra-neuron dependencies.}
    \label{fig:intra_neuron_dependency}
\end{wrapfigure} 

Now we consider the intra-neuron dependency  $\phi^{(l)<2>}_i(t_k, t_m)$ defined between an arbitrary time $t_k$ and a presynaptic firing time  $t_m$ ($t_m < t_k$). From Section \ref{sec:inter_neuron}, the presynpatic firing at time $t_m$ invokes a continuous PSC which has a direct impact on the postsynaptic potential at time $t_k$, which is an inter-neuron dependency. On the other hand, $\phi^{(l)<2>}_i(t_k, t_m)$ corresponds to an indirect effect on the postsynaptic potential via the intra-neuron dependency.

We consider $\phi^{(l)<2>}_i(t_k, t_m)$ specifically under the context of the adopted LIF model, which may occur if the presynaptic neuron spikes at $t_p$ immediately following $t_m$ such that  $t_m<t_p<t_k$. In this case, the presynatpic membrane potential at $t_m$ not only contributes to a PSC  due to the neuron firing, but also affects the membrane potential at the next spike time $t_p$ resulted from the reset  occurring at $t_m$ as described in (\ref{eq:discrete_lif}). The PSC $a_i^{(l)}[t_k]$ has an inter-neuron dependency on membrane potential $u_i^{(l)}[t_p]$ while $u_i^{(l)}[t_p]$ is further affected by the immediately preceding firing time $t_m$ due to the reset of the presynaptic potential at $t_m$.  Recall $s^{(l)}_i[t] = 1$ if neuron $i$ fires at $t$ as in  (\ref{eq:fire_function}).  More precisely, $\phi^{(l)<2>}_i(t_k, t_m)$ takes this indirect intra-neuron effect on $a_i^{(l)}[t_k]$ into consideration %$u_i^{(l)}[t_p]$ 
if  $\exists t_p \in (t_m, t_k) \, \text{such that} \, s^{(l)}_i[t_p] = 1$ and $s^{(l)}_i[t] = 0 \; \forall t \in (t_m, t_p)$, i.e. no other presynaptic spike exists between $t_m$ and $t_p$ 
\begin{equation}
    \label{eq:intra_neuron_dependency}
        \phi^{(l)<2>}_i(t_k, t_m) = \frac{\partial a_i^{(l)}[t_k]}{\partial u_i^{(l)}[t_p]}\frac{\partial u_i^{(l)}[t_p]}{\partial t_m}\frac{\partial t_m}{\partial u_i^{(l)}[t_m]} = \phi_i^{(l)}(t_k, t_p)\frac{\partial (\nu*s^{(l)}_i[t_m])[t_p]}{\partial t_m}\frac{\partial t_m}{\partial u_i^{(l)}[t_m]},
\end{equation}
where $\nu(\cdot)$ is the reset kernel and  $\frac{\partial t_m}{\partial u_i^{(l)}[t_m]}$ is evaluated by (\ref{eq:dt_du}).  In (\ref{eq:intra_neuron_dependency}), $\phi_i^{(l)}(t_k, t_p)$ would have been already computed during the backpropagation process since  $t_p$ is a presynaptic firing time after $t_m$. 

Putting the inter-neuron and intra-neuron dependencies together, one of the key derivatives required in the BP process $\phi_i^{(l)}(t_k, t_m) = \frac{\partial a_i^{(l)}[t_k]}{\partial u_i^{(l)}[t_m]}$ with $t_m<t_p<t_k$ is given by
\small
\begin{equation}
\label{eq:final_rule}
\phi_i^{(l)}(t_k, t_m) = 
\begin{cases}
0   & s^{(l)}_i[t_m] = 0, s_i^{(l)}[t_p] = 0 \; \forall t_p \in (t_m, t_k), \\
\frac{\partial a_i^{(l)}[t_k]}{\partial t_m} \frac{\partial t_m}{\partial u_i^{(l)}[t_m]} & s^{(l)}_i[t_m] = 1, s^{(l)}_i[t_p] = 0 \; \forall t_p \in (t_m, t_k) ,  \\
\frac{\partial a_i^{(l)}[t_k]}{\partial t_m} \frac{\partial t_m}{\partial u_i^{(l)}[t_m]} + \phi^{(l)<2>}_i(t_k, t_m)  & s^{(l)}_i[t_m] = 1, s^{(l)}_i[t_p] = 1, s^{(l)}_i[t] = 0 \; \forall t \in (t_m, t_p),
\end{cases}
\end{equation} 
\normalsize
where $t_p$ is an arbitrary time between $t_m$ and $t_k$, and $\phi^{(l)<2>}_i(t_k, t_m)$ of  (\ref{eq:intra_neuron_dependency})  is considered.  

%To sum it up, 
The complete derivation of TSSL-BP is provided in Section 1 of the Supplementary Material.
There are two key distinctions setting our approach apart from the aforementioned activation smoothing. First, the inter-neuron dependencies are  only considered at pre-synaptic firing times as opposed to all prior time points, latter of which is the case when the activation smoothing is applied with BPTT. The handling adopted in TSSL-BP is a manifestation of the all-or-none firing characteristics of spiking neurons. Second, as  in Figure \ref{fig:inter_neuron_dependency}, the key step in backpropagation is the consideration of the incremental change of spiking times, which is not considered in recent SNNs BP works.

\section{Experiments and Results} \label{sec:results}
We test the proposed TSSL-BP method on four image datasets MNIST~\cite{lecun1998gradient}, N-MNIST~\cite{orchard2015converting}, FashionMNIST~\cite{xiao2017fashion} and CIFAR10~\cite{krizhevsky2014cifar}. We compare TSSL-BP with several state-of-the-art results with the same or similar network sizes including different SNNs BP methods, converted SNNs, and traditional ANNs. The details of practical simulation issues, experimental settings, and datasets preprocessing methods are described in Section 2 of the supplementary material. We have made our Pytorch implementation available online\footnote{\url{https://github.com/stonezwr/TSSL-BP}}. We expect this work would help move the community forward towards enabling high-performance spiking neural networks simulation within short latency.

\subsection{MNIST}
On MNIST~\cite{lecun1998gradient}, we compares the accuracies of the spiking CNNs trained by the TSSL-BP method with ones trained by other algorithms in Table \ref{tb:spikingcnn_mnist}. In our method, the pixel intensities of the image are converted into real-valued spike current to the inputs within a short time window. The proposed TSSL-BP delivers up to $99.53\%$ accuracy and outperforms all other methods except for the ST-RSBP~\cite{zhang2019spike} whose accuracy is slightly higher by  $0.09\%$. However, compared to ST-RSBP, TSSL-BP can train high-performance SNNs with only $5$ time steps, achieving $80\times$ reduction of step count (latency). The accuracy of \cite{zhang2019spike} drops below that of TSSL-BP noticeably under short time windows. In addition, no data augmentation is applied in this experiment, which is adopted in \cite{jin2018hybrid} and \cite{zhang2019spike}. 
%We attribute the efficient learning mainly to two reasons: (1) TSSL-BP can precisely capture the time dependencies and backpropagate the error; (2) Weighted spike trains contains more information and increase the network activity which leads to the fast learning. 

\begin{table*}[ht]
\caption{Performances of Spiking CNNs on MNIST.}
\label{tb:spikingcnn_mnist}
\begin{center}
\begin{small}
\begin{tabular}{lcccccc}
\toprule
Methods & Network & Time steps & Epochs & Mean & Stddev & Best \\
\midrule
Spiking CNN~\cite{lee2016training} & 20C5-P2-50C5-P2-200    & $>200$ & $150$&    &  &  $99.31\%$ \\
STBP~\cite{wu2018spatio} & 15C5-P2-40C5-P2-300 & $>100$ & $200$&  &  & $99.42\%$\\
SLAYER~\cite{shrestha2018slayer} & 12C5-p2-64C5-p2    & $300$ & $100$&  $99.36\%$ & $0.05\%$ &  $99.41\%$  \\
HM2BP~\cite{jin2018hybrid} & 15C5-P2-40C5-P2-300   & $400$ & $100$&  $99.42\%$  & $0.11\%$ &  $99.49\%$       \\
ST-RSBP~\cite{zhang2019spike}  & 15C5-P2-40C5-P2-300   & $400$ & $100$& $99.57\%$ & $0.04\%$ &  $99.62\%$ \\
This work & 15C5-P2-40C5-P2-300     & $\bm{5}$ & $ 100 $ &   $99.50\%$ & $0.02\%$ & $99.53\%$\\
\bottomrule
\multicolumn{7}{p{0.9\linewidth}}{\footnotesize{20C5: convolution layer with 20 of the $5\times5$ filters. P2: pooling layer with $2\times2$ filters.}}
\end{tabular}
\end{small}
\end{center}
\end{table*}

\subsection{N-MNIST}
We test the proposed method on N-MNIST dataset~\cite{orchard2015converting}, the neuromorphic version of the MNIST. The inputs to the networks are spikes rather than real value currents. Table~\ref{tb:nmnist} compares the results obtained by different models on N-MNIST. The SNN trained by our proposed approach naturally processes spatio-temporal spike patterns, achieving the start-of-the-art accuracy of $99.40\%$. It is important to note that our proposed method with the accuracy of $99.28\%$ outperforms the best previously reported results in \cite{shrestha2018slayer}, obtaining $10$ times fewer time steps which leads to significant latency reduction.

\begin{table}[ht]
\caption{Performances on N-MNIST.}
\label{tb:nmnist}
\begin{center}
\begin{small}
\begin{tabular}{lccccccc}
\toprule
Methods & Network & Time steps & Mean & Stddev & Best & Steps reduction\\
\midrule
HM2BP~\cite{jin2018hybrid} & $400-400$ & $600$ & $98.88\%$ & $0.02\%$ & $98.88\%$ & $20$x\\
SLAYER~\cite{shrestha2018slayer}  & $500-500$ & $300$ & $98.89\%$ & $0.06\%$ & $98.95\%$  & $10$x   \\
SLAYER~\cite{shrestha2018slayer}  & 12C5-P2-64C5-P2 & $300$ & $99.20\%$ & $0.02\%$ & $99.22\%$ & $10$x    \\
This work & 12C5-P2-64C5-P2 & $100$ & $\bm{99.35\%}$ & $0.03\%$ & $\bm{99.40\%}$ & $3.3$x\\
This work & 12C5-P2-64C5-P2 & $\bm{30}$ & $99.23\%$ & $0.05\%$ & $99.28\%$ & \\
\bottomrule
\multicolumn{7}{p{0.9\linewidth}}{\footnotesize{All the experiments in this table train the N-MNIST for 100 epochs}}
\end{tabular}
\end{small}
\end{center}
\end{table}

\subsection{FashionMNIST}
We compare several trained fully-connected feedforward SNNs and spiking CNNs on FashionMNIST~\cite{xiao2017fashion}, a more challenging dataset than MNIST.  In Table \ref{tb:fashionmnist},  TSSL-BP achieves $89.80\%$ test accuracy on the fully-connected feedforward network of two hidden layers with each having 400 neurons, outperforming the HM2BP method of \cite{jin2018hybrid}, which is the best previously reported algorithm for feedforward SNNs. TSSL-BP can also deliver the best test accuracy with much fewer time steps. Moreover, TSSL-BP achieves $\bm{92.83\%}$ on the spiking CNN networks, noticeably outperforming the same size non-spiking CNN trained by a standard BP method. 

\begin{table}[ht]
\caption{Performances on FashionMNIST.}
\label{tb:fashionmnist}
\begin{center}
\begin{small}
\begin{tabular}{lcccccc}
\toprule
Methods & Network & Time steps & Epochs & Mean & Stddev & Best \\
\midrule
ANN~\cite{zhang2019spike} & $400-400$ &  & $100$& & & $89.01\%$ \\
HM2BP~\cite{zhang2019spike} & $400-400$ & $400$ & $100$& & & $88.99\%$ \\
This work  & $400-400$ & $\bm{5}$ & $100$ & $89.75\%$ & $0.03\%$ & $\bm{89.80\%}$     \\
\midrule
ANN~\cite{xiao2017fashion} & 32C5-P2-64C5-P2-1024 &  & $100$& & & $91.60\%$ \\
This work & 32C5-P2-64C5-P2-1024 & $5$ & $100$& $92.69\%$ & $0.09\%$ & $\bm{92.83\%}$\\
\bottomrule
\end{tabular}
\end{small}
\end{center}
\end{table}

\subsection{CIFAR10}
Furthermore, we apply the proposed method on the more challenging dataset of CIFAR10~\cite{krizhevsky2014cifar}. As shown in Table \ref{tb:cifar10}, our TSSL-BP method achieves $89.22\%$ accuracy with a mean of $88.98\%$ and a standard deviation of $0.27\%$ under five trials on the first CNN and achieves $91.41\%$ accuracy on the second CNN architecture. TSSL-BP delivers the best result among a previously reported ANN, SNNs converted from pre-trained ANNs, and the spiking CNNs trained by the STBP method of \cite{wu2019direct}. CIFAR10 is a challenging dataset for most of the existing SNNs BP methods since the long latency required by those methods makes them hard to scale to deeper networks. The proposed TSSL-BP not only achieves up to $3.98\%$ accuracy improvement over the work of \cite{wu2019direct} without the additional optimization techniques including neuron normalization and population decoding which are employed in \cite{wu2019direct}, but also utilizes fewer time steps. 

\begin{table}[ht]
\caption{Performances of CNNs on CIFAR10.}
\label{tb:cifar10}
\begin{center}
\begin{small}
\begin{tabular}{lcccc}
\toprule
Methods & Network & Time steps & Epochs & Accuracy \\
\midrule
% ANN\cite{hunsberger2016training} & CNN 1 &  & & $83.72\%$ \\
Converted SNN~\cite{hunsberger2016training}& CNN 1 & $80$ & & $83.52\%$ \\
STBP~\cite{wu2019direct} & CNN 1 &  $8$& $150$& $85.24\%$ \\
This work & CNN 1 & $\bm{5}$& $150$& $\bm{89.22\%}$     \\
\midrule
ANN~\cite{wu2019direct} & CNN 2 &  & & $90.49\%$ \\
Converted SNN~\cite{sengupta2019going} & CNN 2 &  & 200 & $87.46\%$ \\
STBP (without NeuNorm)~\cite{wu2019direct} & CNN 2 &  $8$& $150$& $89.83\%$ \\
STBP (with NeuNorm)~\cite{wu2019direct} & CNN 2 &  $8$& $150$& $90.53\%$ \\
This work & CNN 2 & $\bm{5}$ & $150$& $\bm{91.41\%}$     \\
\bottomrule
\multicolumn{5}{p{0.9\linewidth}}{\footnotesize{CNN 1: 96C3-256C3-P2-384C3-P2-384C3-256C3-1024-1024}} \\
\multicolumn{5}{p{0.9\linewidth}}{\footnotesize{CNN 2: 128C3-256C3-P2-512C3-P2-1024C3-512C3-1024-512}}
\end{tabular}
\end{small}
\end{center}
\end{table}

\subsection{Firing Sparsity}
As presented, the proposed TSSL-BP method can train SNNs with low latency. In the meanwhile, the firing activities in well-trained networks also tend to be sparse. To demonstrate firing sparsity, we select two well-trained SNNs, one for the CIFAR10 and the other for the N-MNIST.

\begin{figure}[ht]
\centering
\begin{minipage}{.45\textwidth}
  \centering
  \includegraphics[width=.95\linewidth]{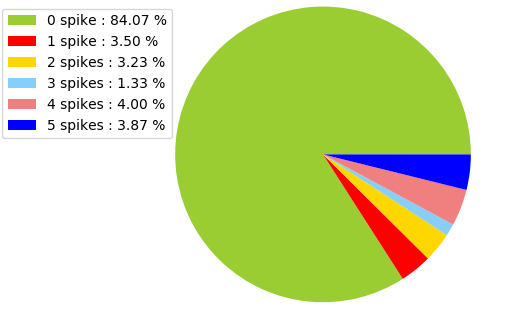}
  \captionof{figure}{Firing activity on CIFAR10}
  \label{fig:cifar10_firing_acitivity}
\end{minipage}%
\begin{minipage}{.45\textwidth}
  \centering
  \includegraphics[width=.95\linewidth]{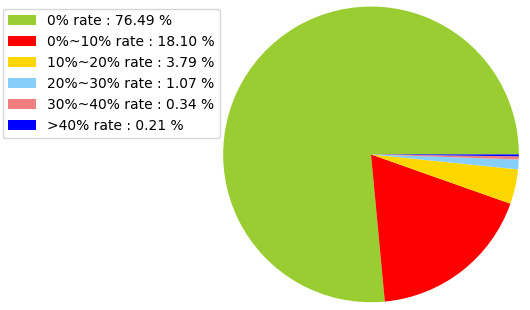}
  \captionof{figure}{Firing activity on N-MNIST}
  \label{fig:nmnist_firing_acitivity}
\end{minipage}
\end{figure}

The CIFAR10 network is simulated over $5$ time steps. We count the percentage of neurons that fire $0, 1, \dots, 5$ times, respectively, and average the percentages over 100 testing samples. As shown in Figure \ref{fig:cifar10_firing_acitivity},  the network's firing activity is sparse. More than $84\%$ of neurons are silent while $12\%$ of neurons fire more than once, and about $4\%$ of neurons fire at every time step. 

The N-MNIST network demonstrated here is simulated over $100$ time steps. The firing rate of each neuron is logged. The number of neurons with a certain range of firing rates is counted and averaged over 100 testing samples. Similarly, as shown in Figure \ref{fig:nmnist_firing_acitivity}, the firing events of the N-MNIST network are also sparse and more than $75\%$ of neurons keep silent. In the meanwhile, there are about $5\%$ of neurons with a firing rate of greater than $10\%$.

\section{Conclusion}
We have presented the novel temporal spike sequence learning via a backpropagation (TSSL-BP) method to train deep SNNs. 
%The main purposes of the proposed TSSL-BP method are to: (1) more precisely handle the non-differentiability of the spiking activation function while keeping all-or-none firing characteristics of spiking neurons; (2) provide a way to train SNNs within a small number of time steps (i.e. latency) to reduce response time, improve accuracy and efficiency. 
% The proposed TSSL-BP method properly computes the error gradient by  capturing inter-neuron and intra-neuron spatiotemporal dependencies. 
Unlike all prior SNNs BP methods, TSSL-BP improves temporal learning precision by circumventing the non-differentiability of the spiking activation function while faithfully reflecting the all-or-none firing characteristics and the underlying structure in the dependencies of spiking neural activities in both space and time.   
%We demonstrate the state-of-the-art  performances in comparison with prior SNN training methods over the MNIST, N-MNIST, FashionMNSIT, and CIFAR10 datasets. 

TSSL-BP provides a universal BP tool for learning arbitrarily specified target firing sequences with high accuracy while achieving low temporal latency. This is in contrast with  most of the existing SNN BP methods which require hundreds of time steps for achieving decent accuracy. The ability in training and inference over a few time steps results in  significant reductions of the computational cost required for training  large/deep SNNs, and the decision time and energy dissipation of the SNN model when deployed on either a general-purpose or a dedicated neurormorphic computing platform. 

%The performances and efficiency achieved by the presented approach may be attributed to the proper handling on spatiotemporal dependencies and precise gradient computation from the all-or-none firing characteristics of spiking neurons. As such, TSSL-BP provides a universal BP tool for learning arbitrarily specified target firing sequences with high accuracy while achieving low temporal latency. It has demonstrated improved scalability to deeper networks and  encouraging results on datasets challenging to spiking network models.

\newpage
% In order to provide a balanced perspective, authors are required to include a statement of the potential broader impact of their work, including its ethical aspects and future societal consequences. Authors should take care to discuss both positive and negative outcomes.
% Authors are required to include a statement of the broader impact of their work, including its ethical aspects and future societal consequences. Authors should discuss both positive and negative outcomes, if any. For instance, authors should discuss a) who may benefit from this research, b) who may be put at disadvantage from this research, c) what are the consequences of failure of the system, and d) whether the task/method leverages biases in the data. If authors believe this is not applicable to them, authors can simply state this.
% Use unnumbered first level headings for this section, which should go at the end of the paper. Note that this section does not count towards the eight pages of content that are allowed.
\section*{Broader Impact}
Our proposed  Temporal Spike Sequence Learning Backpropagation (TSSL-BP) method is able to train deep SNNs while achieving the state-of-the-art efficiency and accuracy.  TSSL-BP breaks down error backpropagation across two types of inter-neuron and intra-neuron dependencies and leads to improved temporal learning precision.  It captures inter-neuron dependencies through presynaptic firing times by considering the all-or-none characteristics of firing activities, and captures intra-neuron dependencies by handling the internal evolution of each neuronal state in time.   

Spiking neural networks offer a very appealing biologically plausible model of computation and may give rise to ultra-low power inference and training on recently emerged large-scale neuromorphic computing hardware.  Due to the difficulties in dealing with the all-or-one characteristics of spiking neurons, however, training of SNNs is a major present challenge and has limited wide adoption of  SNNs models. 

The potential impacts of this work are several-fold:

\textbf{1) Precision}: TSSL-BP offers superior precision in learning arbitrarily specified target temporal sequences, outperforming all recently developed the-state-of-the-art SNN BP methods.

\textbf{2) Low latency}:  TSSL-BP delivers high-precision training over a very short temporal window of a few time steps. This is contrast with many BP methods that require hundreds of time steps for maintaining a decent accuracy. Low latency computation immediately corresponds to fast decision making.

\textbf{3) Scalability and energy efficiency}: The training of SNNs is signficantly more costly than that of the conventional neural networks. The training cost is one major bottleneck to training large/deep SNNs in order to achieve competitive performance. The low latency training capability of TSSL-BP reduces the training cost by more than one order of magnitude and also cuts down the energy dissipation of the training and inference on the deployed computing hardware.

\textbf{4) Community impact}: TSSL-BP has been prototyped based on the widely adopted Pytorch framework and will be made available to the public.  We believe our TSSL-BP code will benefit the brain-inspired computing community from both an algorithmic and neuromorphic computing hardware development perspective. 

\section*{Acknowledgments}
This material is based upon work supported by the National Science Foundation (NSF) under Grants No.1948201 and No.1940761. Any opinions, findings, conclusions or recommendations expressed in this material are those of the authors and do not necessarily reflect the views of NSF, UC Santa Barbara, and their contractors.

\bibliographystyle{plain}

\newpage



\section*{Supplementary material}

## 1.3 Temporal Spike Sequence Learning via Backpropagation (TSSL-BP) Method

We adopt (3) to define the total loss

$$L = \sum_{k=0}^{N_t} E[t_k] = \frac{1}{2}\sum_{k=0}^{N_t}(\boldsymbol{a_d}[t_k] - \boldsymbol{a_s}[t_k])^2. \tag{4}$$

For the neurons in layer $l$, the error gradient with respect to the presynaptic weights matrix $\boldsymbol{W}^{(l)}$ is

$$\frac{\partial L}{\partial \boldsymbol{W}^{(l)}} = \sum_{k=0}^{N_t} \frac{\partial E[t_k]}{\partial \boldsymbol{W}^{(l)}}. \tag{5}$$

(1) reveals that the values of $\boldsymbol{u}^{(l)}$ at time $t_k$ have contribution to all future fires and losses. Using the chain rule, we get

$$\frac{\partial L}{\partial \boldsymbol{W}^{(l)}} = \sum_{k=0}^{N_t}\sum_{m=0}^{k} \frac{\partial E[t_k]}{\partial \boldsymbol{u}^{(l)}[t_m]} \frac{\partial \boldsymbol{u}^{(l)}[t_m]}{\partial \boldsymbol{W}^{(l)}}. \tag{6}$$

By changing the order of summation, (6) can be written as

$$\frac{\partial L}{\partial \boldsymbol{W}^{(l)}} = \sum_{m=0}^{N_t} \frac{\partial \boldsymbol{u}^{(l)}[t_m]}{\partial \boldsymbol{W}^{(l)}} \sum_{k=m}^{N_t} \frac{\partial E[t_k]}{\partial \boldsymbol{u}^{(l)}[t_m]} = \sum_{m=0}^{N_t} \boldsymbol{a}^{(l-1)}[t_m] \sum_{k=m}^{N_t} \frac{\partial E[t_k]}{\partial \boldsymbol{u}^{(l)}[t_m]}. \tag{7}$$

We use $\delta$ to denote the back propagated error at time $t_m$ as $\boldsymbol{\delta}^{(l)}[t_m] = \sum_{k=m}^{N_t} \frac{\partial E[t_k]}{\partial \boldsymbol{u}^{(l)}[t_m]}$.
Therefore, the weights update formula (6) can be written as

$$\frac{\partial L}{\partial \boldsymbol{W}^{(l)}} = \sum_{m=0}^{N_t} \boldsymbol{a}^{(l-1)}[t_m]\boldsymbol{\delta}^{(l)}[t_m]. \tag{8}$$

$\boldsymbol{a}^{(l-1)}[t_m]$ is analogous to the pre-activation in the traditional ANNs which can be easily obtained from (1) in the forward pass. $\boldsymbol{\delta}^{(l)}[t_m]$ is considered in two cases.

**[$l$ is the output layer.]** The $\boldsymbol{\delta}^{(l)}[t_m]$ can be computed from

$$\boldsymbol{\delta}^{(l)}[t_m] = \sum_{k=m}^{N_t} \frac{\partial E[t_k]}{\partial \boldsymbol{a}^{(l)}[t_k]} \frac{\partial \boldsymbol{a}^{(l)}[t_k]}{\partial \boldsymbol{u}^{(l)}[t_m]}. \tag{9}$$

From (4), the first term of (9) is given by

$$\frac{\partial E[t_k]}{\partial \boldsymbol{a}^{(l)}[t_k]} = \frac{1}{2}\frac{\partial(\boldsymbol{a_d}[t_k] - \boldsymbol{a}^{(l)}[t_k])^2}{\partial \boldsymbol{a}^{(l)}[t_k]} = \boldsymbol{a}^{(l)}[t_k] - \boldsymbol{a_d}[t_k]. \tag{10}$$

**[$l$ is a hidden layer.]** $\boldsymbol{\delta}^{(l)}[t_m]$ is derived using the chain rule and (1).

$$\boldsymbol{\delta}^{(l)}[t_m] = \sum_{j=m}^{N_t}\sum_{k=m}^{j} \frac{\partial \boldsymbol{a}^{(l)}[t_k]}{\partial \boldsymbol{u}^{(l)}[t_m]} \left(\frac{\partial \boldsymbol{u}^{(l+1)}[t_k]}{\partial \boldsymbol{a}^{(l)}[t_k]} \frac{\partial E[t_j]}{\partial \boldsymbol{u}^{(l+1)}[t_k]}\right). \tag{11}$$

It is obtained from the fact that membrane potentials $\boldsymbol{u}^{(l)}$ of the neurons in layer $l$ influence their (unweighted) corresponding postsynaptic currents (PSCs) $\boldsymbol{a}^{(l)}$ through fired spikes, and $\boldsymbol{a}^{(l)}$ further affect the membrane potentials $\boldsymbol{u}^{(l+1)}$ in the next layer. By changing the order of summation, maps

the error $\boldsymbol{\delta}$ from layer $l+1$ to layer $l$.

$$\begin{aligned}
\boldsymbol{\delta}^{(l)}[t_m] &= \sum_{k=m}^{N_t} \frac{\partial \boldsymbol{a}^{(l)}[t_k]}{\partial \boldsymbol{u}^{(l)}[t_m]} \sum_{j=k}^{N_t} \frac{\partial \boldsymbol{u}^{(l+1)}[t_k]}{\partial \boldsymbol{a}^{(l)}[t_k]} \frac{\partial E[t_j]}{\partial \boldsymbol{u}^{(l+1)}[t_k]} \\
&= \sum_{k=m}^{N_t} \frac{\partial \boldsymbol{a}^{(l)}[t_k]}{\partial \boldsymbol{u}^{(l)}[t_m]} \sum_{j=k}^{N_t} \boldsymbol{W}^{(l+1)} \frac{\partial E[t_j]}{\partial \boldsymbol{u}^{(l+1)}[t_k]} \\
&= \sum_{k=m}^{N_t} \frac{\partial \boldsymbol{a}^{(l)}[t_k]}{\partial \boldsymbol{u}^{(l)}[t_m]} (\boldsymbol{W}^{(l+1)})^T \boldsymbol{\delta}^{(l+1)}[t_k] \\
&= (\boldsymbol{W}^{(l+1)})^T \sum_{k=m}^{N_t} \frac{\partial \boldsymbol{a}^{(l)}[t_k]}{\partial \boldsymbol{u}^{(l)}[t_m]} \boldsymbol{\delta}^{(l+1)}[t_k].
\end{aligned} \quad (12)$$

The details of the key term $\frac{\partial \boldsymbol{a}^{(l)}[t_k]}{\partial \boldsymbol{u}^{(l)}[t_m]}$ is discussed in Section 3.3.3 and 3.3.4 of the main manuscript. We'll also summarize the derivatives below.

For the key derivative $\frac{\partial a_i^{(l)}[t_k]}{\partial u_i^{(l)}[t_m]}(t_k \geq t_m)$ of each neuron $i$ in layer $l$, we denote $\phi_i^{(l)}(t_k, t_m) = \frac{\partial a_i^{(l)}[t_k]}{\partial u_i^{(l)}[t_m]} = \phi_i^{(l)<1>}(t_k, t_m) + \phi_i^{(l)<2>}(t_k, t_m)$, where $\phi_i^{<1>}(t_k, t_m)$ represents the inter-neuron dependency and $\phi_i^{(l)<2>}(t_k, t_m)$ is the intra-neuron dependency.

Assuming that the presynaptic neuron $i$ spikes at $t_m$, the inter-neuron dependencies can be represented by

$$\phi_i^{(l)<1>}(t_k, t_m) = \frac{\partial a_i^{(l)}[t_k]}{\partial t_m} \frac{\partial t_m}{\partial u_i^{(l)}[t_m]}. \quad (13)$$

From (2) of the main manuscript, the first part of (13) can be calculated as

$$\frac{\partial a_i^{(l)}[t_k]}{\partial t_m} = \frac{\partial (\epsilon * s_i^{(l)}[t_m])[t_k]}{\partial t_m}. \quad (14)$$

We adopt the approach in [1, 3] to compute the second part of (11) as

$$\frac{\partial t_m}{\partial u_i^{(l)}[t_m]} = \frac{-1}{\frac{\partial u_i^{(l)}[t_m]}{\partial t_m}}, \quad (15)$$

where $\frac{\partial u_i^{(l)}[t_m]}{\partial t_m}$ is obtained by differentiating (4). In fact, (15) can be precisely derived on certain conditions.

According to the Figure 6 of the main manuscript, in the LIF model, the intra-neuron dependencies is caused by the firing-and-resetting mechanism. More precisely, $\phi_i^{(l)<2>}(t_k, t_m)$ takes this indirect intra-neuron effect on $a_i^{(l)}[t_k]$ into consideration if $\exists t_p \in (t_m, t_k)$ such that $s_i^{(l)}[t_p] = 1$ and $s_i^{(l)}[t] = 0 \; \forall t \in (t_m, t_p)$, i.e. no other presynaptic spike exists between $t_m$ and $t_k$

$$\begin{aligned}
\phi_i^{<2>}(t_k, t_m) &= \frac{\partial a_i^{(l)}[t_k]}{\partial u_i^{(l)}[t_p]} \frac{\partial u_i^{(l)}[t_p]}{\partial t_m} \frac{\partial t_m}{\partial u_i^{(l)}[t_m]} \\
&= \phi_i(t_k, t_p) \frac{\partial (\nu * s_i^{(l)}[t_m])[t_p]}{\partial t_m} \frac{\partial t_m}{\partial u_i^{(l)}[t_m]},
\end{aligned} \quad (16)$$

where $\nu(\cdot)$ is the reset kernel and $\frac{\partial t_m}{\partial u_i^{(l)}[t_m]}$ is evaluated by (15). In (16), $\phi_i^{(l)}(t_k, t_p)$ would have been already computed during the backpropagation process since $t_p$ is a presynaptic firing time after $t_m$.

To sum it up, we obtain the derivative of loss with respect to weight according to TSSL-BP method as follows:

$$\frac{\partial L}{\partial \boldsymbol{W}^{(l)}} = \sum_{m=0}^{N_t} \boldsymbol{a}^{(l-1)}[t_m]\boldsymbol{\delta}^{(l)}[t_m],$$

$$\boldsymbol{\delta}^{(l)}[t_m] = \begin{cases} \sum_{k=m}^{N_t}(\boldsymbol{a}^{(l)}[t_k] - \boldsymbol{a_d}[t_k])\phi_i^{(l)}(t_k, t_m) & \text{for output layer,} \\ (\boldsymbol{W}^{(l+1)})^T \sum_{k=m}^{N_t} \phi_i^{(l)}(t_k, t_m)\boldsymbol{\delta}^{(l+1)}[t_k] & \text{for hidden layers,} \end{cases}$$

$$\phi_i^{(l)}(t_k, t_m) = \begin{cases} 0 & s_i^{(l)}[t_m] = 0, s_i^{(l)}[t_p] = 0 \; \forall t_p \in (t_m, t_k), \\ \frac{\partial a_i^{(l)}[t_k]}{\partial t_m}\frac{\partial t_m}{\partial u_i^{(l)}[t_m]} & s_i^{(l)}[t_m] = 1, s_i^{(l)}[t_p] = 0 \; \forall t_p \in (t_m, t_k), \\ \frac{\partial a_i^{(l)}[t_k]}{\partial t_m}\frac{\partial t_m}{\partial u_i^{(l)}[t_m]} + \phi_i^{(l)<2>}(t_k, t_m) & s_i^{(l)}[t_m] = 1, \exists t_p \text{ such that } s_i^{(l)}[t_p] = 1, s_i^{(l)}[t] = 0 \; \forall t \in (t_m, t_p), \end{cases}$$

(17)

where $s_i^{(l)}[t]$ is the firing function described in (5) of the main manuscript and $t_p$ is an arbitrary time between $t_m$ and $t_k$.

## 2 Experiments and Results

### 2.1 Experimental Settings

All reported experiments are conducted on an NVIDIA Titan XP GPU. The implementation of TSSL-BP is on the Pytorch framework [9]. The experimented SNNs are based on the LIF model described in (4) of the main manuscript. The simulation step size is set to 1 ms. Only a few time steps are used to demonstrate low-latency spiking neural computation. The parameters like thresholds and learning rates are empirically tuned. Table 1 lists the typical constant values adopted in our experiments. For the time constant, we vary the membrane time constant from 2ms to 16ms. The same performance level has been observed. This indicates that the proposed method can train SNNs with dynamical behaviors across different timescales and the empirically observed results are not very sensitive to the choice of membrane time constant. No axon and synaptic delay or refractory period is used nor is normalization. Dropout is only applied for the experiments on CIFAR10. Adam [4] is adopted as the optimizer. The network models we train or compare with are either fully connected feedforward networks or convolutional neural networks (CNNs). The mean and standard deviation (stddev) of the accuracy reported are obtained by repeating the experiments five times.

Table 1: Parameters settings.

| Parameter | Value | Parameter | Value |
|---|---|---|---|
| simulation step size | 1 ms | Learning Rate $\eta$ | 0.005 |
| Time Constant of Membrane Voltage $\tau_m$ | 4 ms | Time Constant of Synapse $\tau_s$ | 2 ms |
| Threshold $V_{th}$ | 1 mV | Batch Size | 50 |

### 2.2 Input Encoding

For non-spiking datasets such as static images, the most common preprocessing is to use rate coding to convert static real-valued inputs to spiking inputs. However, this requires many time steps for coding the inputs to guarantee good performance. For static images, we directly convert the raw pixel densities into real-valued spike current inputs within a short time window. While for the neuromorphic datasets that originally contain spikes, we still use the spikes as inputs in our experiments.

### 2.3 Handling of Practical Issues

Two practical circumstances need to be taken into consideration as for other spike-time based BP methods like SpikeProp [1, 2]. First, when a spike is produced by the membrane potential $u[t]$ that barely reaches the threshold, the derivative of $u[t]$ w.r.t time is very small. Numerically, this can make (15) large and result in an undesirable large weight update. To mitigate, we set a bound for this

derivative. Second, absence of firing activities in spiking neurons due to low initial weight values block backpropagation through these neurons. We use a warm-up mechanism to bring up the firing activity of the network before applying the BP method. In the warm-up mechanism, we detect the firing events of each neuron. If there's at least one spike within a certain time window, TSSL-BP is applied directly. Otherwise, warm-up is applied, which uses the continuous sigmoid function of membrane potential to approximate the activation function so that the error can be propagated back even when there is no spike.

### 2.4 Selection of the Desired Output Spike Patterns

The desired output spike trains (labels) for different classes are manually selected without much optimization effort. In the experiments with 5 time steps, we set two fixed sequences [0, 1, 0, 1, 1] and [0, 0, 0, 0, 0] where 1 represents a spike and 0 means no spike at a given time step. We adopted a simple scheme: the number of output neurons is same as the number of classes. For each class, the first sequence is chosen to be the target of one (distinct) neuron corresponding to the class, and the second sequence is targeted for all other output neurons.

### 2.5 Datasets

#### 2.5.1 MNIST

The MNIST [6] digit dataset consists of $60,000$ samples for training and $10,000$ for testing, each of which is a $28 \times 28$ grayscale image. Each pixel value of an MNIST image is converted into a real-valued input current. For the fully connected feedforward networks, the inputs are encoded from each $28 \times 28 \times N_t$ image into a 2D $784 \times N_t$ matrix where $N_t$ is the simulation time steps. Each input sample is normalized to the same mean and standard deviation. No data augmentation is applied in our experiments.

#### 2.5.2 N-MNIST

The N-MNIST dataset [8] is a neuromorphic version of the MNIST dataset generated by tilting a Dynamic Version Sensor (DVS) [7] in front of static digit images on a computer monitor. The movement induced pixel intensity changes at each location are encoded as spike trains. Since the intensity can either increase or decrease, two kinds of ON- and OFF-events spike events are recorded. Due to the relative shifts of each image, an image size of $34 \times 34$ is produced. Each sample of the N-MNIST is a spatio-temporal pattern with $34 \times 34 \times 2$ spike sequences lasting for $300ms$ with the resolution of $1us$. It means there are 300000 time steps in the original N-MNIST dataset. In our experiments, we reduce the time resolution of the N-MNIST samples by 3000 times to speed up the simulation. Therefore, the preprocessed samples only have about 100 time steps. We determine that a channel has a spike at a certain time step of the preprocessed sample if there's at least one spike among the corresponding 3000 time steps of the original sample. We demonstrate the result of the preprocessed N-MNIST with 100 time steps in Table 2 of the main manuscript. Moreover, the TSSL-BP method can well train the network only with the first $90ms$ spike sequences of the original dataset which results in 30 time steps after preprocessing. The results are shown in Table 2 of the main manuscript with only 30 time steps which keep the high level of performance as well as significantly reduce the latency.

#### 2.5.3 FashionMNIST

The Fashion-MNIST [11] dataset contains $28 \times 28$ grey-scale images of clothing items, meant to serve as a much more difficult drop-in replacement for the MNIST dataset. The preprocessing steps are the same as MNIST.

#### 2.5.4 CIFAR-10

The CIFAR-10 [5] dataset contains $60,000$ $32 \times 32$ color images in 10 different types of objects. There are $50,000$ training images and $10,000$ testing images. The pixel intensity of each channel is converted into a real-valued input. Similar to what are commonly adopted for preprocessing, the dataset is normalized, and random cropping and horizontal flipping are applied for data augmentation. In addition, dropout layers with a rate of 0.2 are also applied during the training of CIFAR10.



\end{document}